\documentclass[conference]{IEEEtran}
\IEEEoverridecommandlockouts
\usepackage{booktabs}
\usepackage{array}
\usepackage{cite}
\usepackage{amsmath,amssymb,amsfonts}
\usepackage[noend]{algorithmic}
\usepackage{algorithm}
\usepackage{comment}
\usepackage{url}

\usepackage[caption=false, font=footnotesize]{subfig}

\usepackage{graphicx}
\usepackage{textcomp}
\usepackage{xcolor}
\usepackage{hyperref}
\hypersetup{
    colorlinks,
    linkcolor={blue!50!black},
    citecolor={blue!50!black},
    urlcolor={blue!80!black}
}

\def\BibTeX{{\rm B\kern-.05em{\sc i\kern-.025em b}\kern-.08em
    T\kern-.1667em\lower.7ex\hbox{E}\kern-.125emX}}
\begin{document}

\title{Evolutionary scheduling of university activities based on consumption forecasts to minimise electricity costs\\
\thanks{This research was partly funded  by VLAIO project MAMUET (grant number HBC.2018.0529).}
}

\makeatletter
\newcommand{\newlineauthors}{%
  \end{@IEEEauthorhalign}\hfill\mbox{}\par
  \mbox{}\hfill\begin{@IEEEauthorhalign}
}
\makeatother

\author{\IEEEauthorblockN{Julian Ruddick}
\IEEEauthorblockA{\textit{EVERGi, MOBI} \\
\textit{Vrije Universiteit Brussel}\\
Brussels, Belgium \\
julian.jacques.ruddick@vub.be}
\and
\IEEEauthorblockN{Evgenii Genov}
\IEEEauthorblockA{\textit{EVERGi, MOBI} \\
\textit{Vrije Universiteit Brussel}\\
Brussels, Belgium \\
evgenii.genov@vub.be}
\and
\IEEEauthorblockN{Luis Ramirez Camargo}
\IEEEauthorblockA{\textit{EVERGi, MOBI} \\
\textit{Vrije Universiteit Brussel}\\
Brussels, Belgium \\
luis.ramirez.camargo@vub.be}
\newlineauthors
\IEEEauthorblockN{Thierry Coosemans}
\IEEEauthorblockA{\textit{EVERGi, MOBI} \\
\textit{Vrije Universiteit Brussel}\\
Brussels, Belgium \\
thierry.coosemans@vub.be}
\and 
\IEEEauthorblockN{Maarten Messagie}
\IEEEauthorblockA{\textit{EVERGi, MOBI} \\
\textit{Vrije Universiteit Brussel}\\
Brussels, Belgium \\
maarten.messagie@vub.be}

}

\maketitle

\begin{abstract} 

This paper presents a solution to a predict then optimise problem which goal is to reduce the electricity cost of a university campus. The proposed methodology combines a multi-dimensional time series forecast and a novel approach to large-scale optimization. Gradient-boosting method is applied to forecast both generation and consumption time-series of the Monash university campus for the month of November 2020. For the consumption forecasts we employ log transformation to model trend and stabilize variance. Additional seasonality and trend features are added to the model inputs when applicable. The forecasts obtained are used as the base load for the schedule optimisation of university activities and battery usage. The goal of the optimisation is to minimize the electricity cost consisting of the price of electricity and the peak electricity tariff both altered by the load from class activities and battery use as well as the penalty of not scheduling some optional activities. The schedule of the class activities is obtained through evolutionary optimisation using the covariance matrix adaptation evolution strategy and the genetic algorithm. This schedule is then improved through local search by testing possible times for each activity one-by-one. The battery schedule is formulated as a mixed-integer programming problem and solved by the Gurobi solver. This method obtains the second lowest cost when evaluated against 6 other methods presented at an IEEE competition that all used mixed-integer programming and the Gurobi solver to schedule both the activities and the battery use. The code and data used for the paper are publicly available\footnote{\url{https://github.com/EVERGi/predict-then-schedule-university-activities}.}.

\end{abstract}


\section{Introduction}
Whilst the renewable energy source uptake is increasing globally, more flexibility will be needed in electricity sector. Demand response is a source of flexibility projected to play a key role in meeting rising flexibility requirements in the world \cite{InternationalEnergyAgencyIEA2019}. 

Dynamic scheduling of flexible electric load is a method to utilize existing flexibility in the system. The method enables participating in demand response program, while keeping track of an electricity price market and extra spending in peak load tariff. Real-world dynamic problems entail solving an optimization problem bearing partially defined parameters. The missing specifications are recovered using a prediction model. The approach followed at the energy optimization scheduling problem follows the \textit{predict-then-optimize} paradigm.  

An array of methods have been proposed for load scheduling, particularly in energy systems with large flexible loads. Gao et al. \cite{gao2020review} conducts a review on energy-efficient scheduling algorithms in production systems. According to the review, swarm intelligence and evolutionary algorithms are found most applicable to solve energy-efficiency scheduling problems for the large-scale instances. The survey highlights future research directions in use of local-search operators, modelling energy consumption-related constraints and research on applications in specific fields.

Different works have already used evolutionary scheduling on top of forecasted data and obtained good performances in the process. Tušar et al. \cite{tusarevolutionary} compared evolutionary scheduling, a randomized greedy search method and a hybrid between the two to schedule flexible offers for a supply and demand problem based on forecasted electricity loads. The evolutionary scheduling outperformed the two other methods. Ohta et al. \cite{ohtaevolutionary} used an improved multi-objective particle swarm optimization algorithm to schedule air-conditioning temperature based on air-temperature forecast. The method used was shown to be robust to the uncertainty on the air-temperature forecast. Trivedi et al. \cite{trivedievolutionary} used multi-objective evolutionary scheduling to solve a realistic day-ahead thermal generation scheduling problem based on forecasted loads and where unit outage may occur. The authors compares this method with the more commonly used method by system operators to solve the problem as a constrained single-objective optimization problem in deterministic environment. They show that the evolutionary scheduling method obtains better costs and presents a trade-off with emission cost or/and reliability. Mandi et al. \cite{mandi2021predict} views an energy-cost aware scheduling problem as a learning to rank (LTR) problem. The proposed solution uses surrogate loss functions to cache feasible solutions. 

This paper proposes a solution to a real-world problem in energy-cost aware activities scheduling. The problem is first stated within the "IEEE-CIS Technical challenge on Predict+Optimize for renewable energy scheduling" \cite{bergmeir_ieee_2021} and tackles electricity-cost problems of the Monash university. Part of the consumption in the university campus is driven by activities' schedule, which creates potential flexibility in electric demand. The problem is challenging as the solution should combine a multi-dimensional time series forecasting and a large-scale optimization. Therefore, a solution requires expertise in both fields and their interaction. 

Time-series forecasting competitions have long been a principal approach to empirically evaluate several forecasting methods and identify the superior in terms of accuracy performance. Majority of competitive events, however, are largely focused on forecasting and optimization problems in isolation. Previously, there has been only one similar challenge, the 'ICON Challenge on Forecasting and Scheduling' hosted in 2016. The challenge tackled scheduling multiple server jobs while optimizing the energy cost. The challenge  was leaned heavily on the optimization part, with only one variable of electricity price subject to forecasting. The winning solution implements "a constructive heuristics responsible for generating an initial solution and a late acceptance hill climbing algorithm responsible for improving this initial solution" \cite{van2017multi}.

The paper is structured as follows. In Section \ref{sec:statement}, we formulate the scope of the problem, method of evaluation, as well as given conditions and data. The strategy followed while solving the problem is presented in Section \ref{sec:method}. We provide details on the particular workflow, including the methodology in preprocessing, forecasting and scheduling optimization. Section \ref{sec:results} presents an analysis of the results achieved for a prediction accuracy and energy costs optimized via scheduling. Finally, we make the final remarks and present conclusions in Section \ref{sec:conclusions}.

\section{Problem statement}
\label{sec:statement}
The data available from the Monash university in Melbourne includes historical data for energy demand and solar PV production  \cite{bergmeir_ieee_2021}. There is electricity consumption data recorded every 15 minutes from 6 buildings on the Monash Clayton campus, until September 2020. Solar generation data, again with 15 minutes of granularity, sources from 6 rooftop photovoltaic (PV) installations from the Clayton campus, also until September 2020. It also includes energy spot prices for the state of Victoria from the Australian Energy Market Operator \cite{AEMO} as well as weather data from the Australian Bureau of Meteorology \cite{bureau}. Furthermore, ERA5 weather data was supplemented by the competition organizers. The electricity price and weather data are available for the out-of-sample period. Therefore, the final solution is proposed given an assumption of having perfect weather and price forecasts. The problem consists in predicting the energy demand and solar production for the month of November 2020 and schedule the class activities and the battery use for that month in order to minimise the electricity cost.

The schedule cost amounts to energy cost, peak load penalty and subtracts the profit receivable from organizing optional once-off activities. The full formulation of the schedule cost $O$ is shown in (\ref{eq:obj_fun}) where $T$ is the length of the month in 15 minute times steps, $l_t$ is the total power consumption, $e_t$ is the price of electricity, $A$ is the number of once-off activities, $s_a$ is $0$ when the once-off activity $a$ is not scheduled and $1$ when it is scheduled, ${v}_{a}$ is the reward for scheduling once-off activity $a$, $o_a$ is $0$ when once-off activity $a$ is scheduled within working hours and $1$ otherwise and ${p}_{a}$ is the penalty of scheduling once-off activity $a$ outside of working hours.

\begin{equation}
   \begin{aligned}
    O ={ }&\sum_{t=1}^{T} \frac{0.25 l_{t} e_{t}}{1000}+\frac{\left(\max _{t} l_{t}\right)^{2}}{200} -\sum_{a=1}^{A}s_{a} \left(v_{a}-o_{a} p_{a}\right)
    \end{aligned}
    \label{eq:obj_fun}
\end{equation}

The scheduled activities have two types: recurrent and once-off activities. The scheduling of activities needs to follow a set of constraints. The recurrent activities need to be scheduled within working days (Monday-Friday) and working hours (9am-17pm) starting on from the first Monday of the month and occurring at the same day of the week and time for 4 consecutive weeks. The once-off activities can be scheduled outside working hours but get a reduced reward in such case. Recurrent and once-off activities have a set of precedence constraints which are the list of all activities that need to occur at least one day before itself. The activities are defined by its power draw, duration and number of rooms needed to organise the activity. Each building has a number of rooms. Activities need to be scheduled in order that at any time-step the number of rooms occupied by activities does not exceed the number of rooms of all buildings.  

Furthermore, the university has access to two batteries which should be given a schedule that specifies at each time-step if the batteries charge, discharge or do nothing. The effect of the battery schedule on the total power consumption $l_t$ is described in (\ref{eq:load_with_batt}) where $bl_t$ is the power consumption of the buildings, solar panels and scheduled activities, $B$ is the total number of batteries, $m_b$ is the maximum charge and discharge power of battery $b$, $y_b,t$ is $0$ when the battery does not discharge and $1$ when it discharges, $x_b,t$ is $0$ when the battery does not charge and $1$ when the battery charges and $eff_b$ is the efficiency of battery $b$.

\begin{equation}
   \begin{aligned}
    l_t ={ }& bl_{t}+ \sum_{b=1}^{B}m_b\left(y_{b,t}\sqrt{eff_b}-\frac{x_{b,t}}{\sqrt{eff_b}}\right) \quad \forall \medspace t \in T
    \end{aligned}
    \label{eq:load_with_batt}
\end{equation}

The battery can not charge and discharge at the same time $t$ as described in (\ref{eq:charge_discharge}).

\begin{equation}
   \begin{aligned}
    x_{b,t}+y_{b,t} \leq 1 \quad
 \forall \medspace b \in B,\medspace t \in T
    \end{aligned}
    \label{eq:charge_discharge}
\end{equation}

The energy stored in the battery $c_{b,t}$ varies based on the charge and discharge variables $x_b,t$ and $y_b,t$ of the battery as described in (\ref{eq:state_of_charge}) and is bounded between 0 and the maximum capacity $cap_b$ of battery $b$ (see (\ref{eq:soc_bound})). 

\begin{equation}
   \begin{aligned}
    c_{b,t+1} = c_{b,t}+0.25m_b\left(x_{b,t}-y_{b,t}\right)\quad
 \forall \medspace b \in B,\medspace t \in T
    \end{aligned}
    \label{eq:state_of_charge}
\end{equation}

\begin{equation}
   \begin{aligned}
    0 \leq c_{b,t} \leq cap_b \quad
 \forall \medspace b \in B,\medspace t \in T
    \end{aligned}
    \label{eq:soc_bound}
\end{equation}

The given conditions are set as the problem instances. The problem is composed of 5 small instances and 5 large instances. Small instances have 50 recurrent and 20 once-off activities to schedule and large instances 200 recurrent and 100 once-off activities to schedule. An instance codifies a number and parameter description of buildings, batteries and activities to be scheduled. 

\section{Methods}
\label{sec:method}
Due to inherent complexity of both tasks, we address the tasks within two separate workflows. The approach to forecasting and optimization are illustrated in the data flow diagram in Fig.~\ref{fig:flow}. The forecasting method is highlighted in green and the scheduling is in blue. The initial intuition for the forecasting was to use the gradient boosting methods. Large gaps in data and a long forecasting horizon lead to rejecting the idea of using recurrent neural networks and training global models. Furthermore, solutions that employ LightGBM, a popular gradient boosting framework, have dominated in forecasting competitions, particularly the M5 competition \cite{makridakis2020m5}. The main challenge with forecasting is identified at preprocessing the data correctly with an outlook for non-stationarities in data. Efforts were also made to tune the model and select the best features. Concerning scheduling optimization we identified two trajectories for finding solution: a heuristic approach and a constraint programming approach. However, the formulation of the scheduling problem has a high level of complexity, which may not be feasible unless broken down into smaller sub-problems. For the activity scheduling, a base schedule is obtained through evolutionary optimisation. This base schedule is then improved by testing possible times for each activity one-by-one. The battery schedule is formulated as a mixed-integer programming (MIP) problem and solved with the Gurobi solver\cite{gurobi_optimization_gurobi_2021}. 
\begin{figure*}
    \includegraphics[width=\textwidth,height=7.5cm]{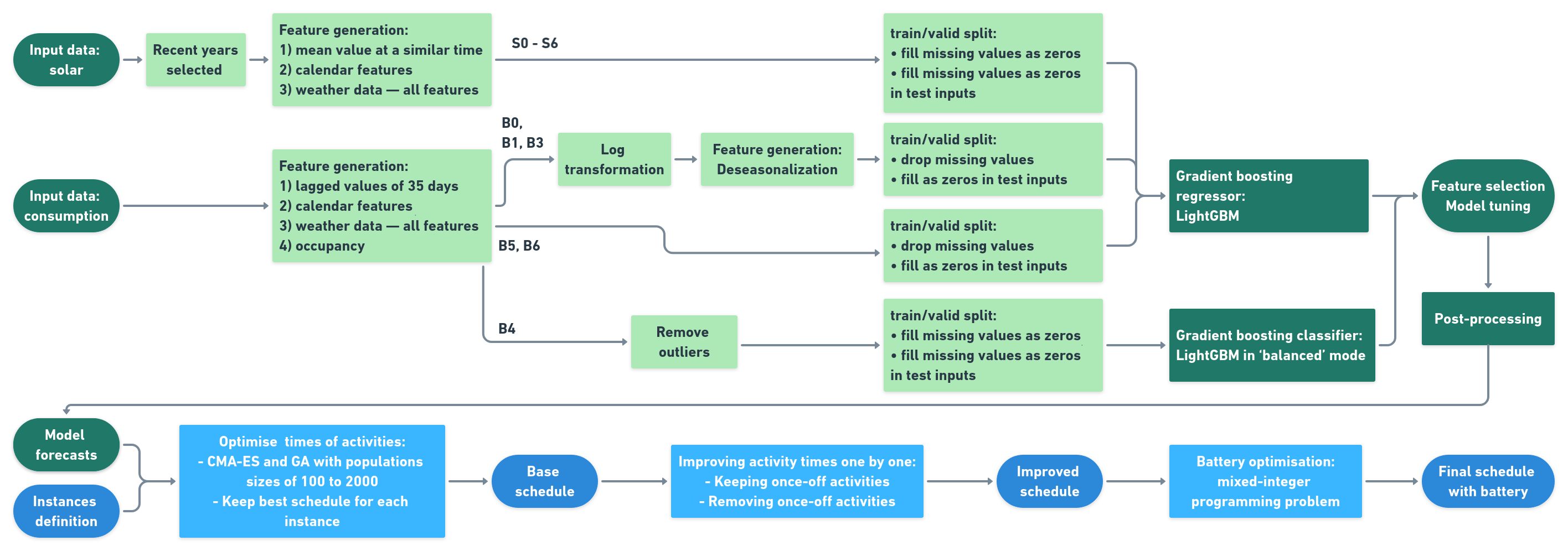}
    \caption{Data flow of forecasting (in green) and schedule optimisation (in blue).}
    \label{fig:flow}
\end{figure*}
\subsection{Forecasting}
\subsubsection{Data pre-processing and feature engineering}
The subject time series record electric consumption in 6 buildings and solar generation in 6 rooftop installation on the Monash Clayton campus. Two of series (Building 0 and Building 3) trace the five-year long period from 2016 up to November 2020, the rest of the series are shorter in length, showing only values for about a year up to November 2020. 

The input data for solar PV generation shows some anomalies, i.e. long patches of zero measurements and changes in trend, particularly at the start of recording. For training the models, the early weeks are discarded from the solar series

Calendar features are generated to aid the learning procedure of seasonal patterns for energy consumption in buildings. The features are time of day, weekday and time of year. The weekday feature is encoded as a categorical feature. Others are encoded with sin and cos functions in order to capture the cyclicity with their scale. A working day feature is created. Some information is provided on the used capacity during periods of the academic calendar and restrictions in response to COVID-19 pandemic. We incorporate occupancy fractions in buildings as a separate feature. The occupancy in earlier dates is estimated as monthly consumption values re-scaled to a $[0, 1]$ range with a min-max normalization. 

Both for the generation and consumption data, the input time series were split into training and validation sets. The validation set is similar duration to the test, to be long enough to ensure robustness to weekly seasonalities.  A validation partition is used to tune hyper-parameters, select features and compare the methods' performance. The validation set is selected to start at fixed origin, as the test set is released at once. The last observed values for the month-long period, prior to the forecasted month, are selected to be the validation set for all time series except for Building 5. A significant data drift, i.e. change in data distribution, is observed for consumption in October in the building 5 time-series. A prior month is set for validation.

We calculate the trend and seasonality components for all points of the consumption. The values for these components are also projected into the forecasted period. This is done using the $prophet$ forecasting library \cite{taylor2018forecasting}. The underlying mode performs a multi-seasonal additive decomposition. Back-testing it showed that using the additional features improves final prediction in all buildings but 5 and 6. Feature selection is handled while tuning the model. Pairwise correlation is calculated using the Pearson correlation method. Features that exceed the correlation coefficient threshold, which is subject to hyperparameter optimization, are discarded. In most cases, feature selection shows to be insignificant to the final accuracy on the validation data. 
\subsubsection{Model implementation}
We utilize LightGBM \cite{ke2017lightgbm}, a gradient boosting method for solving non-linear regression and classification problems, to forecast the time-series for the month ahead. The objective function is set to minimize mean absolute error (MAE). MAE is proportionate to Mean Absolute Scaled Error (MASE). We tune the model for hyperparameters and feature selection on the validation set. 

The ranges for hyperparameters are based off default values provided in package's documentation, adjusted for the size of the series. The values are provided in Table \ref{fig:hyperp}.
\begin{table}[htbp]
\caption{Initial hyperparameter ranges}
\centering
\begin{tabular}{|l|l|}
\hline
\textbf{method}               & {[}'gbdt', 'dart', 'goss'{]} \\ \hline
\textbf{learning rate}        & 0.01 - 0.3                   \\ \hline
\textbf{number of  leaves}    & 20-3\,000                      \\ \hline
\textbf{maximum depth}        & {[}-1, 5, 10, 15, 20, 25{]}  \\ \hline
\textbf{minimum data in leaf} & 20-100                       \\ \hline
\textbf{lambda l1}            & 0-100                        \\ \hline
\textbf{lambda l2}            & 0-100                        \\ \hline
\textbf{correlation value}    & 0.6 - 1                      \\ \hline
\end{tabular}
\label{fig:hyperp}
\end{table}

Building 4 time series stands out due to a strictly discrete distribution of consumption values and a high number of missing values. We approach forecasting this series as the  multi-class classification problem with an unbalanced dataset. The model is set to 'balanced' mode, where weights are adjusted inversely proportional to class frequencies in the training data

\subsection{Schedule optimisation}

Before the start of the optimisation, a $precedence\ level$ is calculated for each activity. The $precedence\ level$ of an activity is the minimum number of days necessary before the activity to be able to satisfy the precedence constraints of all the activities. A $level\ after$ value is also calculated for each activity. This value is the minimum number of days needed after the activity to be able to satisfy the precedence constraints of all the activities. 

\subsubsection{Base schedule} 

The base activity schedule is obtained through the optimisation of an evolutionary algorithm. Two different evolutionary algorithms were tested, the Covariance matrix adaptation evolution strategy (CMA-ES) \cite{hansen_reducing_2003} and the genetic algorithm \cite{goldberg_genetic_2006}. The evolution process starts by creating a population of possible schedules. Each individual has two components for each recurrent activity and one component for each once-off activity. The components for the recurrent activities are the day and the time of the day at which the activity should be scheduled. For each activity the days which can be selected depend on how many activity levels are below and above it in the precedence directed graph. The days which can be selected are the five days of the week from which the first $x$ days and the $y$ last days are removed, where $x$ is the $precedence\ level$ of the activity and $y$ is the $levels\ after$ value of the activity. By doing this we discard the days for which it is impossible to schedule the recurrent activity due to the precedence constraints. The time of the day for the recurrent activities is selected from the start of the working day to the end of the working day minus the duration of the activity. The time for the once-off activities is selected from the first time of the month to the last time of the month minus the duration of the activity (Alg. \ref{alg:base} line \ref{alg:gen_pop}).

To satisfy the precedence constraints for the recurrent and once-off activities, the days of the activities are changed gradually starting from the activities with the lowest precedence level. If an activity has a precedence scheduled for the same or a later day, the activity is rescheduled to the day after the latest day of its precedence activities. The time of the day stays the same for the rescheduled recurrent and once-off activities. If due to this process a once-off activity is scheduled after the last day of the month, this activity is discarded (Alg. \ref{alg:base} line \ref{alg:enf_prec}).

When the precedence constraint has been satisfied, the room constraints need to be enforced. Rooms are assigned first for the recurrent activities and second for the once-off activities using the same process. Activities for which the product of the duration and the number of rooms occupied by the activity is the highest are the first to be assigned to a room. An activity is assigned to rooms available with the lowest ids first. If there are not sufficient rooms available to fit the activity, the activity is scheduled at a different time of the day and during working hours for the recurrent activities. This process tests times gradually from closest to furthest to original given time. If no rooms are available in the selected day, the solution is discarded by giving it a high score of \$200\,000 (Alg. \ref{alg:base} line \ref{alg:def_rooms}).

Once-off activities which increase the score of the objective function from the previously obtained feasible schedule are removed. For each once-off activity, the electricity cost of running the activity, the value and the penalty of itself and all the activities necessary to schedule this activity are summed to give the benefit of scheduling the activity. This benefit defines the impact of the once-off activities has on the objective function without taking the electricity consumption peak cost into account. The more negative this benefit is, the more it will decrease the objective function. The activity with the largest negative benefit and all the activities necessary to schedule this activity are kept for the final solution. The same process is repeatedly executed with the electricity cost, the activity value and the penalty of the kept activities set to 0 until the reward of all remaining activities are positive, in which case the remaining activities are discarded. This process removes all once-off activities, which increase the objective function score. Some kept once-off activities may still increase the objective function score via the peak cost (Alg. \ref{alg:base} line \ref{alg:remove_o-o}).

The obtained schedule is evaluated through the objective function and the total cost is given to the evolutionary algorithm (Alg. \ref{alg:base} line \ref{alg:obj_func}). Moreover, the evolutionary algorithm optimises the selected days and times of the recurrent and once-off activities to minimise the objective function scores of the obtained schedules (Alg. \ref{alg:base} lines \ref{alg:evol_pop} and \ref{alg:gen_pop}).

\begin{algorithm}
\footnotesize{\scriptsize}
\caption{Obtain base schedule}
\label{alg:base}
 \begin{algorithmic}[1]
 \renewcommand{\algorithmicrequire}{\textbf{Input:}}
 \renewcommand{\algorithmicensure}{\textbf{Output:}}
 \REQUIRE instance
 \ENSURE  base schedule
  \WHILE {stop criteria is false}
  \STATE Evolutionary algorithm generate population within possible times \label{alg:gen_pop}
  \FOR {individual in new population}
  \STATE Enforce precedence constraints and define day and time of all activities \label{alg:enf_prec}
  \STATE Define rooms for all activities \label{alg:def_rooms}
  \STATE Remove once-off activities with negative impact on electricity cost\label{alg:remove_o-o}
  \STATE Calculate electricity cost \label{alg:obj_func}
  \ENDFOR
  \STATE Evolve population based on previous individuals and associated electricity costs \label{alg:evol_pop}
  \ENDWHILE 
 \RETURN base schedule
 \end{algorithmic} 

\end{algorithm}

\subsubsection{Improved activity schedule}

From the base schedule, an improved schedule is obtained by modifying the time of the activities one by one. This improvement is done in two versions, one improving the base schedule and a second improving the base schedule from which the once-off activities are removed (Alg. \ref{alg:impr} line \ref{alg:remove_o-o_2}). The improvement starts by modifying the times of the recurrent activities then the once-off activities (Alg. \ref{alg:impr} line \ref{alg:rec_then_o-o}). The time of the activities are modified one by one keeping the time of the other activities fixed and starting with the activities with the lowest $precedence\ level$ (Alg. \ref{alg:impr} lines \ref{alg:days_in_month} and \ref{alg:day_equal_prec}). Once a better time for an activity is found, the better time is given to that activity and the process continues with this new improved solution. In a first phase, the activities try all times within the days that allow all activities to be scheduled and that respect the precedence constraints of the activities already in the schedule (Alg. \ref{alg:impr} line \ref{alg:recom_days}). In a second phase, the activities try all times that respect the precedence constraints of the activities already in the schedule (Alg. \ref{alg:impr} line \ref{alg:all_days}). This means that in the second phase the once-off activities can try times that discard other once-off activities to be scheduled due to precedence constraints. Overall this process finds better times for the once-off and recurrent activities and schedules new once-off activities that were previously not scheduled.

\begin{algorithm}
\caption{Improve base schedule}
\footnotesize{\scriptsize}
\label{alg:impr}
 \begin{algorithmic}[1]
 \renewcommand{\algorithmicrequire}{\textbf{Input:}}
 \renewcommand{\algorithmicensure}{\textbf{Output:}}
 \REQUIRE base schedule
 \ENSURE improved schedule
 \IF {remove once off = true}
 \STATE remove once off activities from base schedule \label{alg:remove_o-o_2}
 \ENDIF
 \FOR {$i:=1$ to $2$}
  \FOR {$activity$ in all activities starting with recurrent} \label{alg:rec_then_o-o}
  \FOR{$day$ in days of the month} \label{alg:days_in_month}
  \IF {$activity$ precedence level = $day$} \label{alg:day_equal_prec}
  \IF {$i=1$}
  \STATE improve time of $activity$ within recommended days \label{alg:recom_days}
  \ELSIF{$i=2$}
  \STATE improve time of $activity$ within all times \label{alg:all_days}
  \ENDIF
  \ENDIF
  \ENDFOR
  \ENDFOR
  \ENDFOR
 \RETURN improved schedule
 \end{algorithmic} 
 
 \end{algorithm}

\subsubsection{Battery schedule}

The schedule of the batteries is found with the activity times of the improved schedule. The behaviour of the batteries described in (\ref{eq:load_with_batt}) to (\ref{eq:state_of_charge}) and the objective function (\ref{eq:obj_fun}) are modelled as a mixed-integer problem. The Gurobi solver \cite{gurobi_optimization_gurobi_2021} was used to minimise the objective function with $x_b,t$ and $y_b,t$ as the variables that describe the battery schedule. The $lb_t$ value in (\ref{eq:load_with_batt}) is here the power consumption of the improved schedule. This process calculates the schedule of the battery for the given improved schedule with imperfect forecast of the buildings and solar panels loads.

\section{Results and discussion}
\label{sec:results}
\subsection{Forecasting}
Table \ref{tab:fcst} provides the accuracy scores achieved with the final model configuration. The accuracy is assessed with the MASE, which is defined in (\ref{eq:MASE}).
\begin{equation}
M A S E=\frac{\sum_{k=M+1}^{M+h}\left|F_{k}-Y_{k}\right|}{\frac{h}{M-S} \sum_{k=S+1}^{M}\left|Y_{k}-Y_{k-S}\right|}
\label{eq:MASE}
\end{equation}
After validating the model using the October data, we proceed to apply the same model to forecast the subsequent month, November 2020. The model generally gives predictions with MASE smaller than 1. Therefore, it  succeeds at beating a seasonal persistence of previous 28 days. The LGB model is more consistent at accurate prediction of photovoltaic generation sequences. 
The aggregate error of forecasts in all buildings and solar installations is visualized in the Fig.~\ref{fig:forecast_agg}. The time period shown in the figure examines the aggregate consumption from Monday 23 of November 2020 to the end of the scheduling time for readability reasons. A significant underestimation is observed at the peak values. Such behaviour is assumed to be caused by models failing to detect a change in trend occurring between the months of October and November. The LGB model are tuned using the October data for validation, thus overfitting to the patterns observed in that particular period.
\begin{table}[htbp]
\caption{Forecasting accuracy scores for October and November}
\centering
\begin{tabular}{ @{} lll @{} }
\toprule
          & \multicolumn{2}{c}{MASE}  \\ 
          & October  & November  \\ \midrule
Building0  & 0.422 & 1.201    \\
Building1  & 0.695 & 1.134    \\
Building3  & 0.932 & 0.640     \\
Building4  & 1.271 & 0.810   \\
Building5  & 0.171 & 0.949   \\
Building6  & 0.948 & 1.018      \\
Solar0 & 0.895 & 1.044  \\
Solar1     & 0.325  & 0.399  \\
Solar3     & 0.859 & 0.525  \\
Solar2     & 0.424 & 0.722   \\
Solar5     & 0.554 & 0.517  \\
Solar4     & 0.446 & 0.729  \\
\midrule
mean       & 0.662 & 0.807   \\ \bottomrule
\end{tabular}
\label{tab:fcst}
\end{table}
\begin{figure}[htbp]
\centerline{\includegraphics[width=\linewidth]{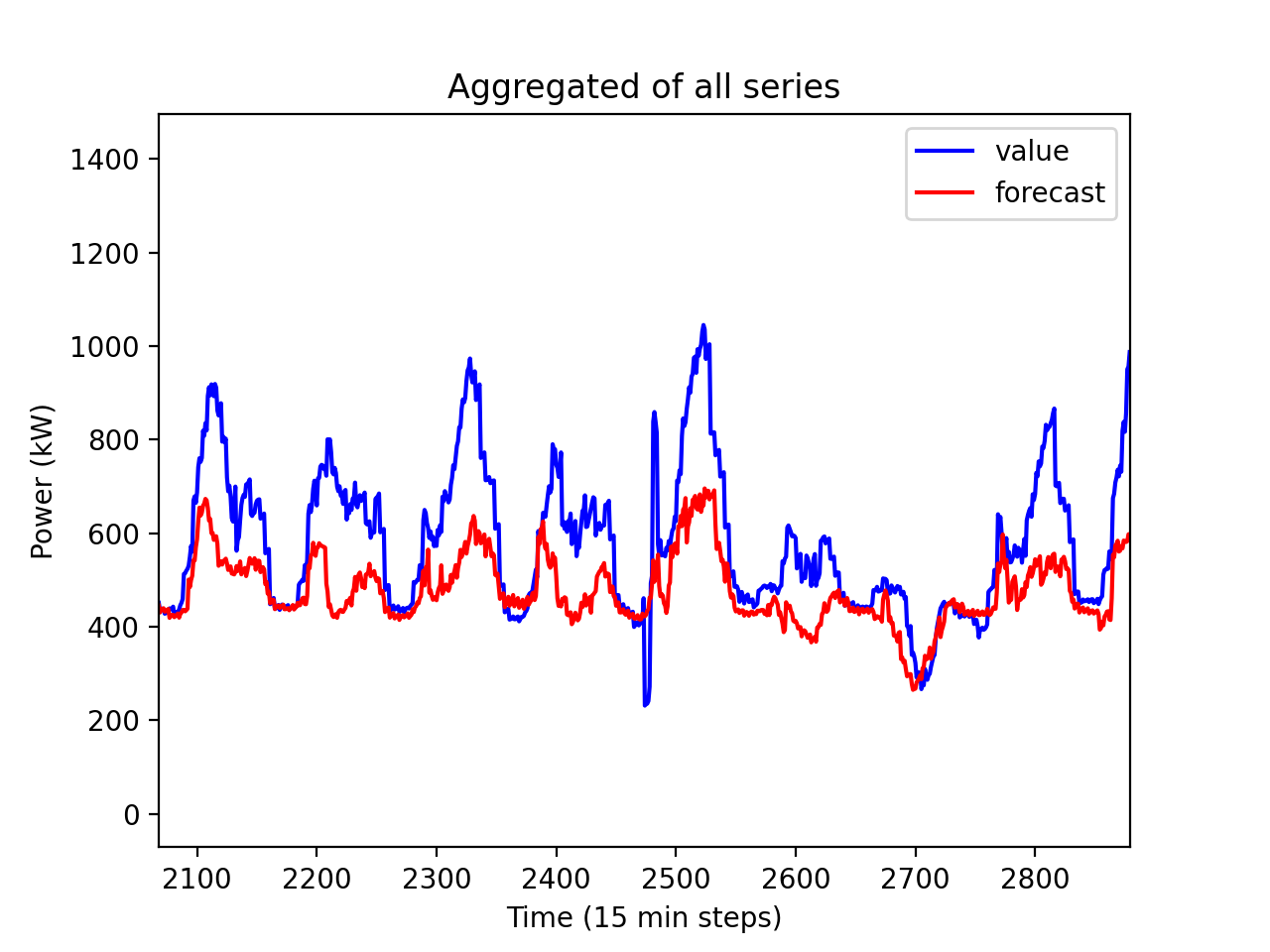}}
\caption{Comparison of actual aggregated load vs the forecasted values from Monday 23 of November 2020 to the end of the scheduling time. In red is the forecasted load of the buildings minus the production of the solar panels. The blue line indicates the actual load.}
\label{fig:forecast_agg}
\end{figure}

\subsection{Schedule optimisation}

\subsubsection{Base schedule analysis}
The GA and CMA-ES were tested with population sizes from 100 to 2\,000 (see Fig.~\ref{fig:base_res}). The stopping criteria for CMA-ES is met when the $f$ tolerance is smaller than 100 or the x tolerance is smaller than 1. The stopping criteria for the GA is met when no improvement larger than 1 is found for 500 generations in a row. Both stopping criteria were tuned to stop the evolution process when no or very slight improvements were found for multiple generations. CMA-ES was implemented using the pygmo library \cite{Biscani2020} and starts with a $\sigma$ of 0.5. The GA uses steady-state selection as selection operator with a selection rate of 10\% of the parent population, the crossover operator is single-point crossover, the mutation operator is random resetting with a mutation probability of 10\%.  The selection rate of the parent population was chosen, all other parameters and operators are the default ones found in the PyGAD library \cite{gad_pygad_2021}.

For small instances CMA-ES and the GA both get close or exceed with all populations size the base schedule used for the competition submission (see Fig. \ref{fig:base_res}). However for the large instances, CMA-ES outperforms the GA for all population sizes except 2\,000. The base schedule used for the competition submission had 3 small schedules obtained with the GA and the 7 others obtained with CMA-ES. 

\begin{figure}[htbp]
\centering
  \subfloat[Small instances\label{fig:long_small}]{
       \includegraphics[width=\linewidth]{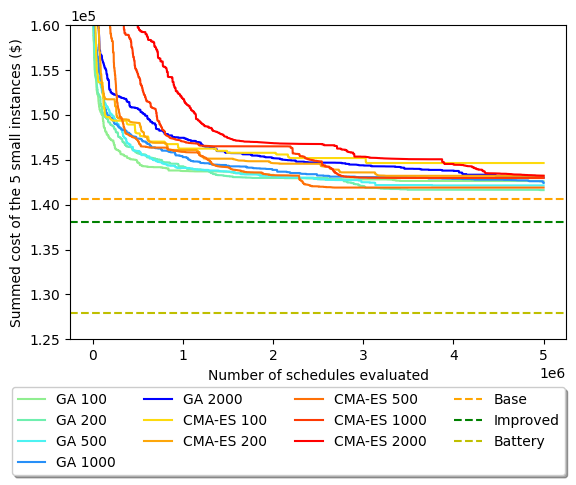}}
    \hfill
    \subfloat[Large instances\label{fig:long_large}]{
       \includegraphics[width=\linewidth]{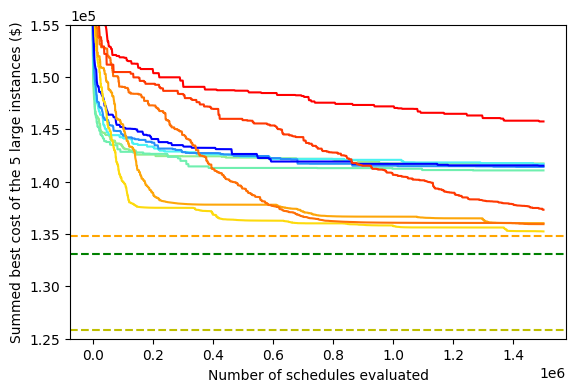}}
    \hfill

\caption{Visualisation of the optimisation process to find a base schedule with the GA and CMA-ES approach for different population sizes. Each solid line represents the summed best cost obtained for the 5 instances of each category (small in Fig. \ref{fig:long_small} and large in Fig. \ref{fig:long_large}) during consecutive optimisation processes. Once the evolutionary process reached its stopping criteria, a new evolutionary process for the same instance is started. The horizontal dashed lines are the costs of the schedules that obtained the lowest cost for each instance during all the runs and at the three phases of the scheduling process.}
\label{fig:base_res}
\end{figure}
\subsubsection{Improvement analysis}

All improvement methods for the activity schedule reduce the final cost (see Fig.~\ref{fig:impro}). The methods where the once-off activities are removed seem to have a slightly better average but the method where the once-off activities are kept does yield in some cases better results. The improved schedule used for the competition submission had 4 schedules obtained by keeping the once-off activities and 6 obtained by removing the once-off activities.

Fig.~\ref{fig:impro} shows that the battery schedule decreases the final score more than the schedule improvement. For small instances the interquartile range of the boxplot is large while for large instances the interquartile range is small. A small interquartile range would indicate that whatever activity schedule is given, the improvement from the battery is always approximately the same and therefore that separating the activity and the battery scheduling is reasonable. The results seem to indicate that this is true for large instances but not necessarily for small instances.

\begin{figure}[htbp]
\centerline{\includegraphics[width=\linewidth]{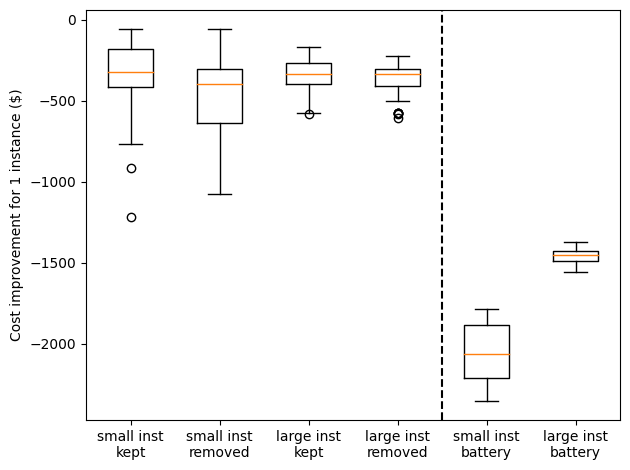}}
\caption{Box-plots of the different methods used to improve the base schedule and the battery schedule. The box-plots for the improvement methods are generated on 40 values each, which are the costs of the improvement of the 8 best base solution found for each instance. The box plots for the battery schedule contain 80 values each, which are the cost obtained by adding the battery schedule to the 160 improved schedules from the 4 other box-plots.}
\label{fig:impro}
\end{figure}

\subsubsection{Overall performance} \label{sec:overall_perf}

The best schedules found with this method have a total electricity cost of \$253\,691.95 when evaluated with the forecasted loads and a cost of \$332\,740.74 when using the real data for November 2020. The difference in cost is caused by the underestimation of forecasted electricity consumption during working hours as shown in Fig.~\ref{fig:forecast_agg}. This is the second lowest cost when evaluated against 6 other methods presented at an IEEE competition \cite{bergmeir_ieee_2021} that all a MIP and the Gurobi solver to schedule both the activities and the battery use \cite{DBLP:journals/corr/abs-2112-03595,beanmethodology,limmerfinal,yuan2021optimal,stratigakos:hal-03449920}. 

This method is exactly the same that obtained the fourth best cost at this IEEE competition but for which a bug preventing the schedule of one of the two provided batteries was discovered and corrected. 

\subsubsection{Example schedule and forecast}\label{sec:example}

Fig.~\ref{fig:example} shows the best activity schedule found for the small\_0 instance. Only the schedule from Monday 23 of November 2020 to the end of the scheduling time are shown for readability reasons. The recurrent activities seem well scheduled to keep a low max peak power value. The once-off activities are placed only during working hours and multiple are scheduled in the last two days which have low electricity price and no recurrent activity. There is nonetheless a visible improvement that can be made to this schedule, 7 out of the 20 once-off activities have not yet been scheduled and between time-steps 2\,676 and 2\,709 the prices are negative. Shifting once-off activities from the 2nd last day to this time-slot and adding non-scheduled once-off activities to the last two working days decreases the cost. By doing this step manually the cost for this instance was improved from \$28\,482.20 to \$28\,335.73. This indicates that the schedule optimisation has still room for improvement regarding once-off activities.

\subsubsection{Time evaluation}

A complete scheduling process was executed logging the time of the three scheduling phases. The whole process was executed on a laptop computer with a 12 thread Intel\textregistered \  Core\textcopyright \ i7-9850H CPU and 32 GB DDR4 RAM. 

The base schedule phase was executed in parallel on the 10 instances of the problem and stopped after 12 hours. The algorithm used was CMA-ES and a population of 100 was selected. During this time an average of 975\,100 and 3\,914\,480 base schedules were created and evaluated for each of the 5 large instances and 5 small instances respectively. 

The 10 best base schedules per instance were then improved in parallel trying both base schedules with once-off activities removed and kept. The improvement process executed in 4 minutes for schedules from small instances and 14 minutes from schedules from large instances. 

The battery schedule was calculated for the 10 best improved schedules. This process was done sequentially and with a time limited of 20 minutes for each schedule to avoid the RAM used to exceed 32 GB. The optimisation took a total time of 1 hour and 42 minutes with all schedules from large instances hitting the 20 minute limit and schedules from small instances being solved each in 22 seconds on average. 

The whole process was executed in 13 hours and 56 minutes and obtained a cost of \$335\,136.88 when evaluating with the real data of November 2020. This cost is close to the cost obtained in section \ref{sec:overall_perf} where both the GA and CMA-ES were used with different population sizes.

\begin{figure}[htbp]
\centerline{\includegraphics[width=\linewidth]{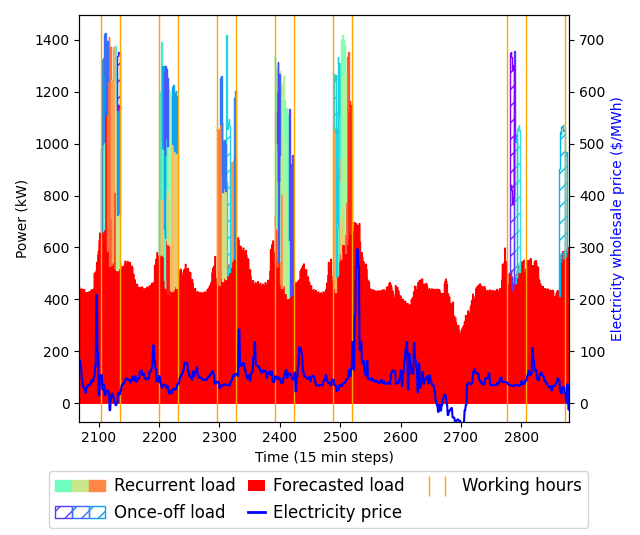}}
\caption{Representation of the best found improved schedule for the small\_0 instance. In red is the forecasted load of the buildings minus the production of the solar panels. The load from the recurrent activities are represented in different colors for each activity. The orange vertical lines represent the start and the end of the working hours of the problem. The blue line is the price of electricity.}
\label{fig:example}
\end{figure}

\subsubsection{Other methods presented at the competition}

For the competition\cite{bergmeir_ieee_2021}, the seven best methods including this one were selected for presentation. 
For the forecasting top 4 solutions use applications of a gradient forest model or gradient boosted trees. The approaches mainly differ in data preprocessing steps and feature engineering. This includes using additional lags for weather, removing outliers and selecting the size of the training data. Remarkably, the best prediction used a naive median prediction for two out of 6 buildings.
For the scheduling, all six other methods solved the problem as a MIP and used the Gurobi solver for both activity and battery schedule. The method in first place optimised the schedule over six different forecasts which seems to be the main advantage over our method \cite{DBLP:journals/corr/abs-2112-03595}.

\section{Conclusion}
\label{sec:conclusions}
We solve a 'predict then optimise' problem within the premises of a university campus. Energy consumption and generation of the university buildings and solar panels are forecasted and used to schedule class activities and the use of two batteries.

The forecasting problem is approached with a deterministic method using seasonal and trend  decomposition followed by a gradient boosting model. Generated seasonality and trend components are used as exogenous inputs. The combination of time series decomposition and gradient boosting seems to be effective. 

To obtain the base schedule, similar performances are achieved with CMA-ES and the GA for small instances but for large instances CMA-ES performs better. The improved schedules obtained through iterations reduces the cost for electricity of the campus and schedules more once-off activities. This improved schedule can still be improved as is shown by the example in section \ref{sec:example}. Ideally the two steps to obtain the base and the improve schedules could be merged into one evolutionary scheduling step that includes more once-off activities. The battery schedule decreases effectively the cost. Scheduling batteries and activities together instead of sequentially could yield lower costs.

\bibliographystyle{unsrt} 
\bibliography{Tech_challenge} 

\end{document}